\documentclass[10pt,twocolumn,letterpaper]{article}

\usepackage[pagenumbers]{cvpr} 
\usepackage{lineno}

\usepackage[dvipsnames]{xcolor}
\usepackage{caption}
\usepackage{placeins}
\definecolor{cvprblue}{rgb}{0.21,0.49,0.74}
\usepackage[pagebackref,breaklinks,colorlinks,citecolor=cvprblue]{hyperref}

\def\paperID{162} 
\def\confName{3DV\xspace}
\def\confYear{2026\xspace}

\title{CAOA - Completion-Assisted Object-CAD Alignment}

\author{Hiranya Garbha Kumar\\
University at Albany\\
Albany, NY, USA\\
{\tt\small hgkumar@albany.edu}
\and
Minhas Kamal\\
University at Albany\\
Albany, NY, USA\\
{\tt\small mxkamal@albany.edu}
\and
Balakrishnan Prabhakaran\\
University at Albany\\
Albany, NY, USA\\
{\tt\small bprabhakaran@albany.edu}
}

\begin{document}
\maketitle
\begin{abstract}
\vspace{-2mm}
{Accurately aligning CAD models to their corresponding objects in indoor RGB-D scans is a central challenge in 3D semantic reconstruction. The task requires estimating a 9-Degree-of-Freedom (DoF) pose—position, rotation, and scale along three axes—but is hindered by noisy and incomplete scans, as well as segmentation errors that cause geometric distortions.
We present Completion-Assisted Object-CAD Alignment (CAOA), a method that integrates a semantically and contextually aware point cloud completion module with a symmetry-aware relative pose estimation algorithm, enabling precise alignment of CAD models to scanned objects. 
Existing completion methods are typically trained and evaluated on synthetic datasets, which often fail to generalize to real-world scans. To bridge this gap, we introduce a synthetic data generation strategy tailored to indoor scenes, significantly reducing the synthetic-to-real domain gap—validated through quantitative comparisons with widely used completion datasets. In addition, we release S2C-Completion, an expert-annotated dataset of over 8,500 object-CAD pairs from Scan2CAD, created for real-world indoor single-object completion and intended as a new benchmark for this task.
For object-CAD alignment, we incorporate symmetry information via a symmetry-aware loss, improving robustness to symmetric ambiguities. On the Scan2CAD benchmark, CAOA achieves a 17\% accuracy improvement over state-of-the-art methods.
All code, datasets, and annotation tools will be publicly available on \href{https://github.com/kumarhiranya/S2CCompletion}{GitHub}.}
\end{abstract}    
\vspace{-5mm}
\section{Introduction}
\vspace{-2mm}
\label{sec:intro}
\begin{figure*}[t]
    \centering
    \includegraphics[width=0.95\textwidth]{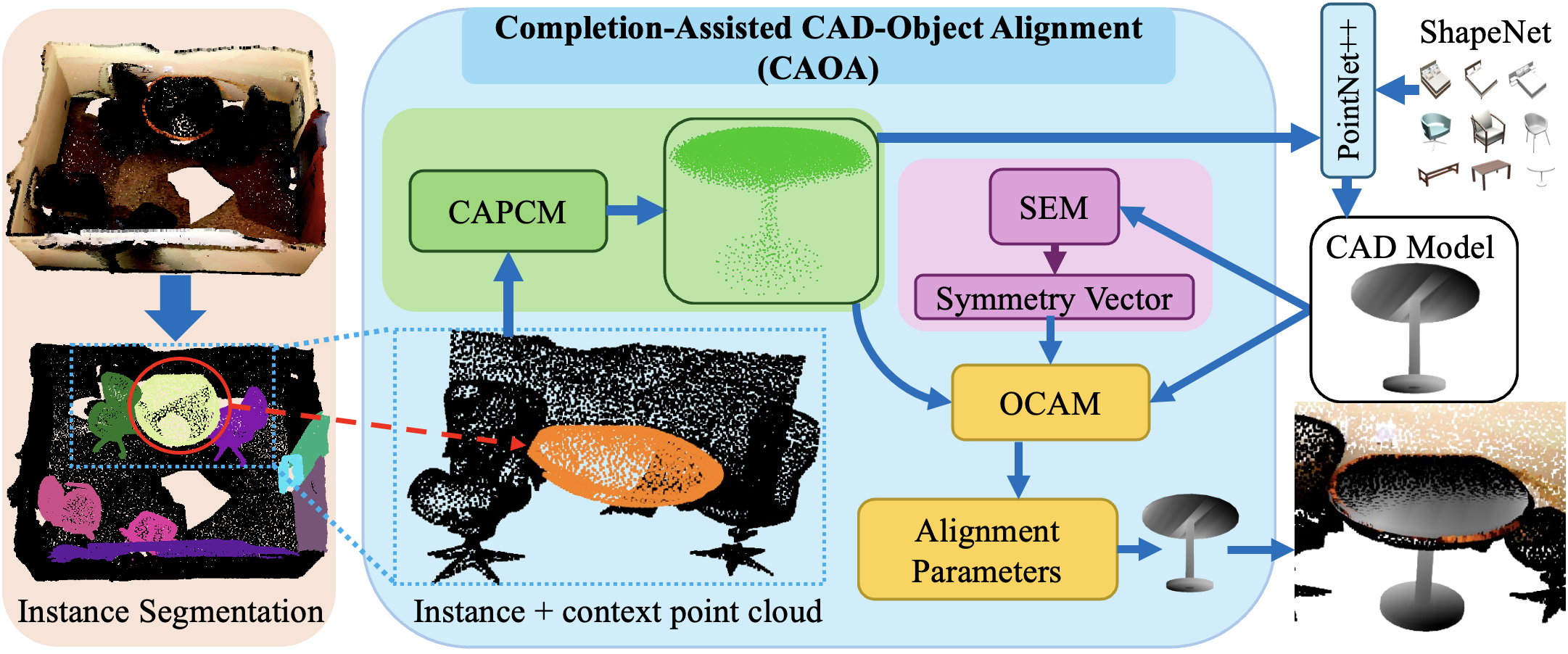} 
    \caption{
    Overview of CAOA: The input to CAOA is a 3D room scan and its corresponding instance segmentation mask. Using this mask, object point clouds (red circle) and surrounding context points (blue dotted rectangle) are extracted from the scan. These are processed by the Context-Aware Point Cloud Completion Module (CAPCM) to produce a completed point cloud (green). Uniformly sampled CAD model points are passed through the Symmetry Encoder Module (SEM) to generate a symmetry vector. The completed point cloud, CAD point cloud, and symmetry vector are then used by the Object–CAD Alignment Module (OCAM) to estimate the final alignment parameters.
    }
    \label{fig:overview}
\end{figure*}

Recent indoor 3D semantic reconstruction methods \cite{avetisyan_scan2cad_2019, avetisyan_scenecad_2020, avetisyan_end--end_2019} integrate high-level semantic information with geometric data, enabling the generation of more complete and interpretable models compared to traditional mesh-based surface reconstruction techniques \cite{dai_bundlefusion_2017, siddiqui_retrievalfuse_2021, kulkarni_whats_2022}. These semantic approaches effectively address challenges such as occlusions and incomplete data, while also producing lightweight representations that are well-suited for a wide range of downstream applications, such as creating interactive virtual environments, digital twinning, etc.
RGB-D scans are effective for indoor 3D semantic reconstruction due to their comprehensive representation of environments. However, challenges like clutter, occlusion, and unreliable depth sensing result in noisy, incomplete scans. Traditional post-processing methods often address these issues but can introduce artifacts and lose detail.
These challenges are critical for tasks associated with 3D semantic reconstruction, such as CAD model retrieval, object-CAD alignment, and object-based CAD texturing.
In this work, we focus on enhancing object-CAD alignment within semantic reconstruction for indoor scenes.

The quality of indoor scans presents significant challenges for accurate CAD alignment, as noise, incompleteness, and errors from earlier steps, like 3D object segmentation, can distort an object's geometry, complicating pose estimation. Current methods, such as energy optimization \cite{avetisyan_scan2cad_2019}, loss functions \cite{avetisyan_end--end_2019, kumar2024cis2vr}, and object-layout optimization \cite{avetisyan_scenecad_2020}, aim to improve alignment, but their effectiveness is limited by poor scan quality.


\vspace{-2mm}
\subsection{Proposed Approach}
\vspace{-2mm}
To address the aforementioned challenges in object-CAD alignment, we introduce Completion-Assisted Object-CAD Alignment (CAOA), a novel approach that leverages point cloud completion to enhance alignment accuracy. An overview of the approach is shown in Figure \ref{fig:overview}.
CAOA consists of 3 modules:
\begin{itemize}
    \item \textbf{Context-Aware Point Cloud Completion Module (CAPCM)}: 
    CAOA processes incomplete and noisy object point clouds using a CAPCM, which generates cleaner, more complete representations. CAPCM is trained with S2C-Completion, a new expert-annotated dataset for real-world indoor object point cloud completion, enabling context-aware training. Additionally, we augment the training with a new synthetic dataset, ShapeNet-Indoor(SN-Indoor), generated from the ShapeNet dataset using novel techniques tailored to indoor environments.
    \item \textbf{Symmetry Encoder Module (SEM)}: To further improve alignment accuracy, we incorporate SEM, which encodes symmetry information through a 3D Transformer-based feature encoder \cite{wu2024point}. 
    \item \textbf{Object-CAD Alignment Module (OCAM)}: The output of CAPCM—a more complete and accurate representation of real-world objects—is combined with the symmetry vector from SEM and a matching CAD model. These 3 form the input for the OCAM, which estimates the final alignment parameters. 
\end{itemize}
By addressing the limitations of incomplete and noisy point clouds through completion and leveraging symmetry, our approach significantly enhances the robustness and accuracy of object-CAD alignment.
In summary, the primary contributions of this work are as follows:
\begin{itemize} 
    \item A novel pose estimation approach that incorporates context-aware point cloud completion to mitigate issues related to noise, incompleteness, and segmentation errors in object instances. 
    \item We introduce S2C-Completion, a real-world indoor object point cloud completion dataset of over 8,500 object-CAD pairings derived from Scan2CAD\cite{avetisyan_scan2cad_2019} to train and benchmark point cloud completion algorithms. 
    \item We introduce the SN-Indoor dataset, derived from ShapeNet \cite{chang2015shapenet}, and generated using synthetic data techniques specifically designed for indoor environments. This approach enables better generalization to real-world data. Additionally, we assess the generalizability of existing synthetic datasets on S2C-Completion by evaluating them with a leading point cloud completion algorithm.
    \item A training methodology leveraging a symmetry encoder and a symmetry-aware loss formulation to learn robust pose features. 
\end{itemize}


\vspace{-2mm}
\section{Related Work}
\label{sec:related}
\vspace{-2mm}
In this section, we discuss existing works in the domain on semantic reconstruction and datasets for point-cloud completion and object-CAD alignment. Semantic reconstruction approaches generally fall into two categories: modular, multi-step methods \cite{kumar2024cis2vr, ainetter2023automatically} and unified, end-to-end methods \cite{avetisyan_end--end_2019} \cite{avetisyan_scenecad_2020} \cite{langer2024FastCADRC}. The flexibility of multi-step methods allows them to integrate state-of-the-art methods optimized for specific tasks, such as segmentation or object completion or pose estimation, thereby benefiting from the latest advances in these areas. In contrast, end-to-end methods are custom-built and trained holistically for the task, enabling the model to learn the complete data transformation pipeline and potentially achieve faster processing times.

\vspace{-2mm}
\subsection{Semantic Reconstruction}
\vspace{-2mm}
Semantic reconstruction typically involves CAD retrieval, object-CAD alignment and layout estimation. In the domain of CAD retrieval, early methods such as the one proposed by Li et al. \cite{li2015cgf12573} relied on using handcrafted features, often derived from local histograms. 
Among modern methods employing learned feature descriptors, 3DMatch \cite{zeng_3dmatch_2017} employs a Siamese network to extract discriminative features common to both scan and CAD objects.
Recent work on deformation-based CAD retrieval \cite{uy2021joint, di_u-red_2023, zhang_kp-red_2024} has substantially enhanced the geometric fidelity of semantic reconstructions. By enabling part-wise deformation of CAD meshes to more closely conform to scan objects, these approaches achieve high-quality reconstruction even when CAD model resources are sparse or retrieval quality is limited. However, it is important to note that while most deformation-based methods integrate a CAD retrieval algorithm, they assume an existing object–CAD alignment as a prerequisite, rather than eliminating the need for such alignment.

Among methods for object-CAD alignment, Scan2CAD \cite{avetisyan_scan2cad_2019}, introduces learnable parameters to establish correspondences between CAD models and scan objects, using an iterative energy optimization algorithm for alignment. The authors in \cite{avetisyan_end--end_2019} further developed an end-to-end pipeline for processing 3D scans to produce aligned CAD models, integrating segmentation, layout detection, and CAD object retrieval and alignment into a single, efficient process. SceneCAD \cite{avetisyan_scenecad_2020} approaches CAD alignment from a global perspective, jointly considering object arrangement and scene layout. This enables SceneCAD to leverage contextual information, achieving globally consistent alignment by exploiting inter-object relationships within the scene. 
Ainetter et al. \cite{ainetter2023automatically} propose an unsupervised iterative approach to aligning CAD models to their scan counterparts.
In CIS2VR \cite{kumar2024cis2vr}, the authors take a modular approach to semantic reconstruction by decoupling 3D object segmentation from pose estimation. It utilizes object point clouds extracted through 3D instance segmentation of scans to infer 9-DoF poses and align CAD models.

\vspace{-2mm}
\subsection{Datasets}
\label{related-pccompletion}
\vspace{-2mm}
\paragraph{Point Cloud Completion} Although numerous studies have introduced point cloud completion algorithms for both scenes~\cite{stutz_learning_2020, wang_deep_2020, song_semantic_2017} and objects~\cite{mittal_autosdf_2023, dai_shape_2017}, including recent advancements~\cite{sun2025dynamic, misik2023cocca, salihu2024deepspf}, these approaches predominantly depend on synthetic datasets for training and quantitative evaluation.
The primary challenge for synthetic datasets, in the context of indoor settings, is emulating characteristics of real-world point clouds, such as noise and irregularities introduced by sensors, and occlusion caused by limited perspectives and environmental clutter.
Synthetic datasets such as PCN \cite{yuan2018pcn} and ShapeNet-55/34 \cite{yu2021pointr}, derived from the ShapeNet \cite{chang2015shapenet} dataset, try to address these challenges by using various methods such as perspective projection and point cloud cropping.
The KITTI \cite{geiger2013kitti} dataset, in contrast, contains partial point clouds obtained from real-world scans. However, it lacks ground truth data, making suitable only for qualitative analysis.

\vspace{-4mm}
\paragraph{Object-CAD Alignment} While several datasets provide CAD model alignment data for 2D images, including IKEA objects \cite{lim2013IKEA}, Pix3D \cite{sun2018pix3d}, and PASCAL 3D+ \cite{xiang2014pascal}, these only include annotations for 6-DoF alignment. Currently, Scan2CAD \cite{avetisyan_scan2cad_2019} is the only large-scale dataset dedicated to object-CAD alignment in indoor 3D scans. It provides a set of aligned ShapeNet \cite{chang2015shapenet} CAD models for each scan in ScanNet while using 9-DoF alignment parameters.

\begin{figure}[htbp]
    \centering
    \includegraphics[width=\columnwidth]{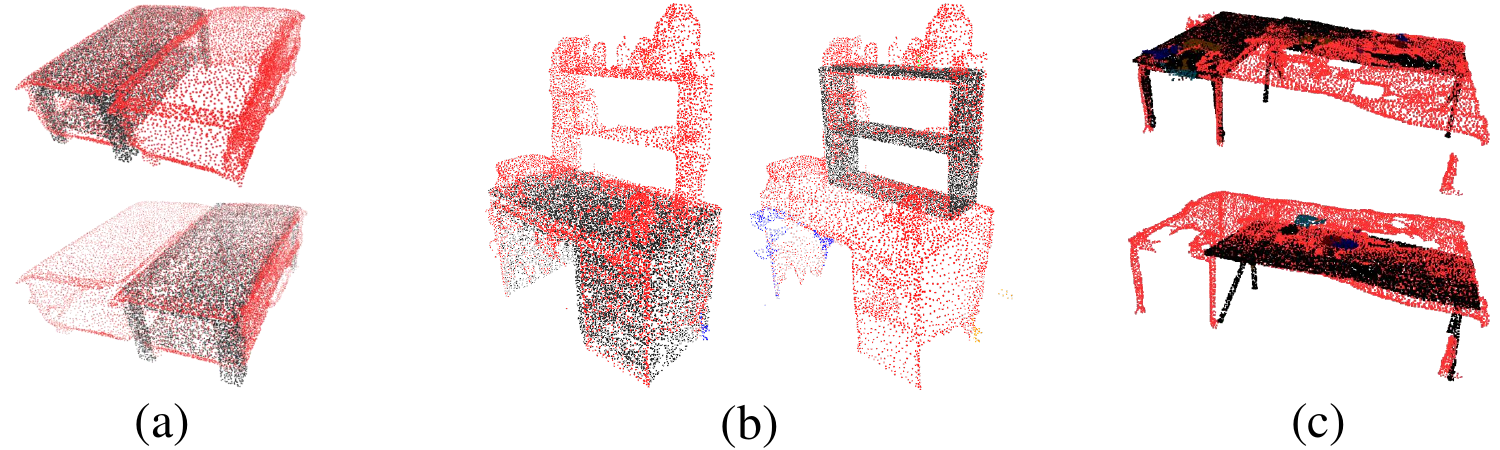}
    \caption{Annotations from Scan2CAD with CAD model (black) and corresponding ScanNet ground truth instance (red) point clouds. Each pair shows the discrepancy in ground truth annotations between Scan2CAD's aligned CAD models and ScanNet's object instances.}
    \label{fig:s2c-annos}
\end{figure}

\vspace{-3mm}
\paragraph{Research Gap} 
Despite advancements in object-CAD alignment algorithms, the poor quality of scanned point cloud objects pose a significant challenge. While point cloud completion algorithms can help address this issue, they are primarily trained on synthetic data and a substantial gap persists between the quality of synthetic and real-world point clouds. Consequently, algorithms trained exclusively on synthetic data struggle to generalize effectively to real-world data, as demonstrated by our experiments detailed in \ref{exp-odg-datasets}. Additionally, in the context of learning object-CAD alignment, Scan2CAD, designed for end-to-end optimization, lacks explicit annotations linking each CAD model and alignment parameter to individual ScanNet ground truth instances. This makes it unsuitable for training algorithms focused on object point clouds. While some methods have used intersection-over-union (IoU) thresholds to address this, the discrepancy between Scan2CAD and ScanNet ground truths (illustrated in Figure \ref{fig:s2c-annos}), coupled with the incomplete nature of object point clouds leading to low IoUs even for correct matches, makes it difficult to handle using such approaches, often causing algorithms to learn sub-optimal features.

\vspace{-4mm}
\section{Completion-Assisted Object-CAD Alignment (CAOA)}
\label{sec:method}
The complete CAOA pipeline is shown in Figure~\ref{fig:overview}. Starting from an indoor scan as a point cloud, we extract object point clouds using an off-the-shelf 3D instance segmentation algorithm~\cite{shin2024spherical}. This modular design enables CAOA to adopt advances in 3D instance segmentation without retraining the entire pipeline. The following subsections describe CAOA's modules.

\vspace{-2mm}
\subsection{Context-Aware Point Cloud Completion (CAPCM)}
\label{completion-assisted-pose-est}
\begin{figure}[htbp]
    \centering
    \includegraphics[width=\columnwidth]{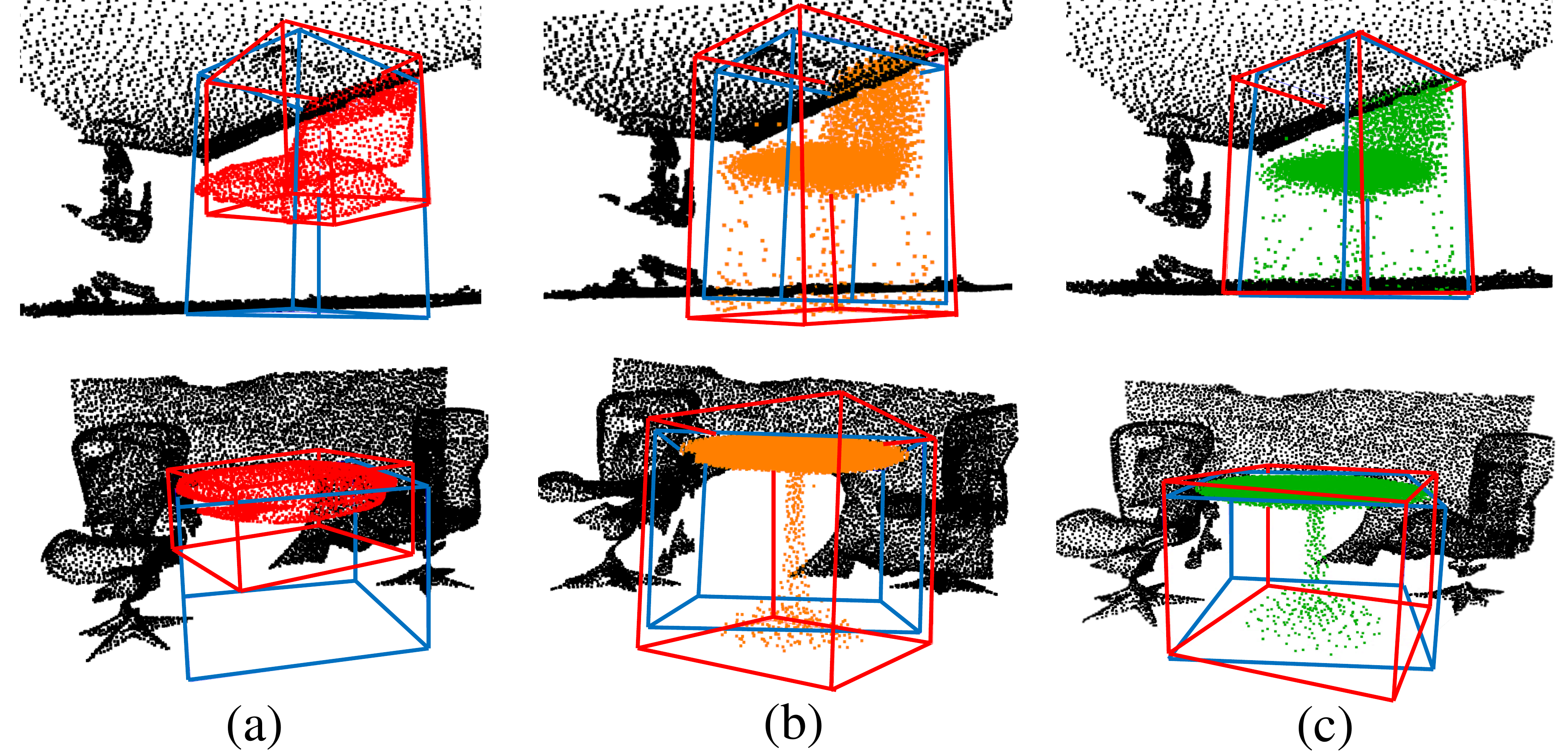}
    \caption{Comparison of pose estimation on raw instance point cloud (a), completion without context (b), context-aware completion (c), with context points shown in black. Ground truth and predicted poses are visualized using blue and red bounding boxes, respectively. Note that although context points are shown for all, they are used only in the context-aware (c) setting.}
    \label{fig:completion-comparison}
\end{figure}

\vspace{-2mm}
\paragraph{Point Cloud Completion}
As shown in Figure \ref{fig:completion-comparison}(a), object point clouds from indoor scans are often noisy and incomplete, with missing features that can significantly alter the object's geometry and affect object-CAD alignment. These issues can lead to slower convergence and suboptimal performance during training.
To address this, CAPCM leverages point cloud completion algorithms. However, due to the lack of real-world datasets for object point cloud completion, these methods are typically trained and validated on synthetic data, which often does not generalize well to real-world scenarios. Moreover, most synthetic datasets rely on generic techniques like perspective projection and linear cropping, which fail to accurately replicate indoor scans.

To facilitate the training and benchmarking of point cloud completion methods on real-world indoor scenes, we introduce the S2C-Completion dataset, specifically designed for indoor environments. In addition, we present SN-Indoor, a new synthetic dataset that incorporates techniques optimized for simulating indoor settings, enhancing the generalization of models trained on real-world indoor data. Detailed descriptions of both datasets are provided in Sections \ref{S2C-Completion-dataset} and \ref{shapenet-indoor}. To demonstrate the effectiveness of these datasets, we evaluate them by training and testing a state-of-the-art point cloud completion algorithm proposed by Cai et al. \cite{cai2024odg}, as detailed in Section \ref{exp-odg-datasets}.

\vspace{-3mm}
\paragraph{Context-Aware Setting}
In cases where large parts of an object are missing, we observe that point cloud completion algorithms trained on pairs of incomplete and complete objects tend to over or under generate point clouds, with the completed point clouds ending up with dimensions that are inconsistent with their environments (as shown in Figure \ref{fig:completion-comparison}).
To address this issue, CAPCM incorporates a context-aware approach for point cloud completion: in addition to the incomplete object point cloud, we incorporate point clouds from surrounding objects or structures within a defined context radius. This additional contextual information enhances the algorithm's understanding of the surrounding scene, leading to results that are more consistent with the environment.


To extract context points $P_{ctxt}$ for a given object point cloud $P_{obj}$ within a scene point cloud $P_{scene}$, we begin by creating an axis-aligned bounding box (AABB) $BB_{obj}$ around the object.
With a specified context radius $R_{ctxt}$, we define a new context bounding box ($BB_{ctxt}$), sharing the same center as $BB_{obj}$, with dimensions adjusted as follows:
\begin{align}
Dim_{ctxt} = Dim_{obj} + R_{ctxt}
\end{align}
Where $Dim_{ctxt}$ and $Dim_{obj}$ are dimensions of the context and object bounding boxes respectively. Using $BB_{ctxt}$, we extract $P_{ctxt}$ from $P_{scene}$ by extracting all points within this bounding box, excluding points corresponding to the object $P_{obj}$.
To help the algorithm distinguish between the object ($P_{obj}$) and context ($P_{ctxt}$) point clouds, we append a 1 to the spatial features of $P_{obj}$ and a -1 to those of $P_{ctxt}$:
\begin{align}
F_{obj} = (X_{obj}, Y_{obj}, Z_{obj}, 1) \\
F_{ctxt} = (X_{ctxt}, Y_{ctxt}, Z_{ctxt}, -1) \\
F_{in} = F_{obj} \oplus F_{ctxt}; P_{in} = P_{obj} \oplus P_{ctxt}
\end{align}
Here, $F_{obj}$ and $F_{ctxt}$ represent the spatial features of the object and context point clouds, $\oplus$ denotes the concatenation operator, while $P_{in}$ and $F_{in}$ denote the final input 3D coordinates and features, respectively. 

For training in the normal setting (without context data), the input consists of the incomplete object point cloud $P_{obj}$ and features $F_{obj}$, both represented by 3D coordinates $(X_{obj}, Y_{obj}, Z_{obj})$. The target point cloud $P_{trgt}$ is obtained by uniformly sampling points from the aligned CAD model’s mesh, with coordinates $(X_{trgt}, Y_{trgt}, Z_{trgt})$. In the context-aware scenario, the inputs to the algorithm are $F_{in}$ and $P_{in}$, while the target point cloud $P_{trgt}$ remains unchanged. We compare the outputs from both settings in Figure \ref{fig:completion-comparison}, where the context-aware approach demonstrates improved consistency in the generated point cloud relative to its surrounding context. Empirical studies indicated an optimal value of  \( R_{ctxt} = 100 \, \text{cm} \), as increasing the radius further had no impact on the overall performance of CAOA. Further details on training the CAPCM are discussed in \ref{cad-alignment-exp}.

{We then use PointNet-based shape descriptors \cite{qi2017pointnet++} derived from the completed object point cloud to retrieve a CAD model from the ShapeNet dataset \cite{chang2015shapenet} that is both geometrically and semantically similar.
}


\begin{figure}[htbp]
    \centering
    \includegraphics[width=\columnwidth]{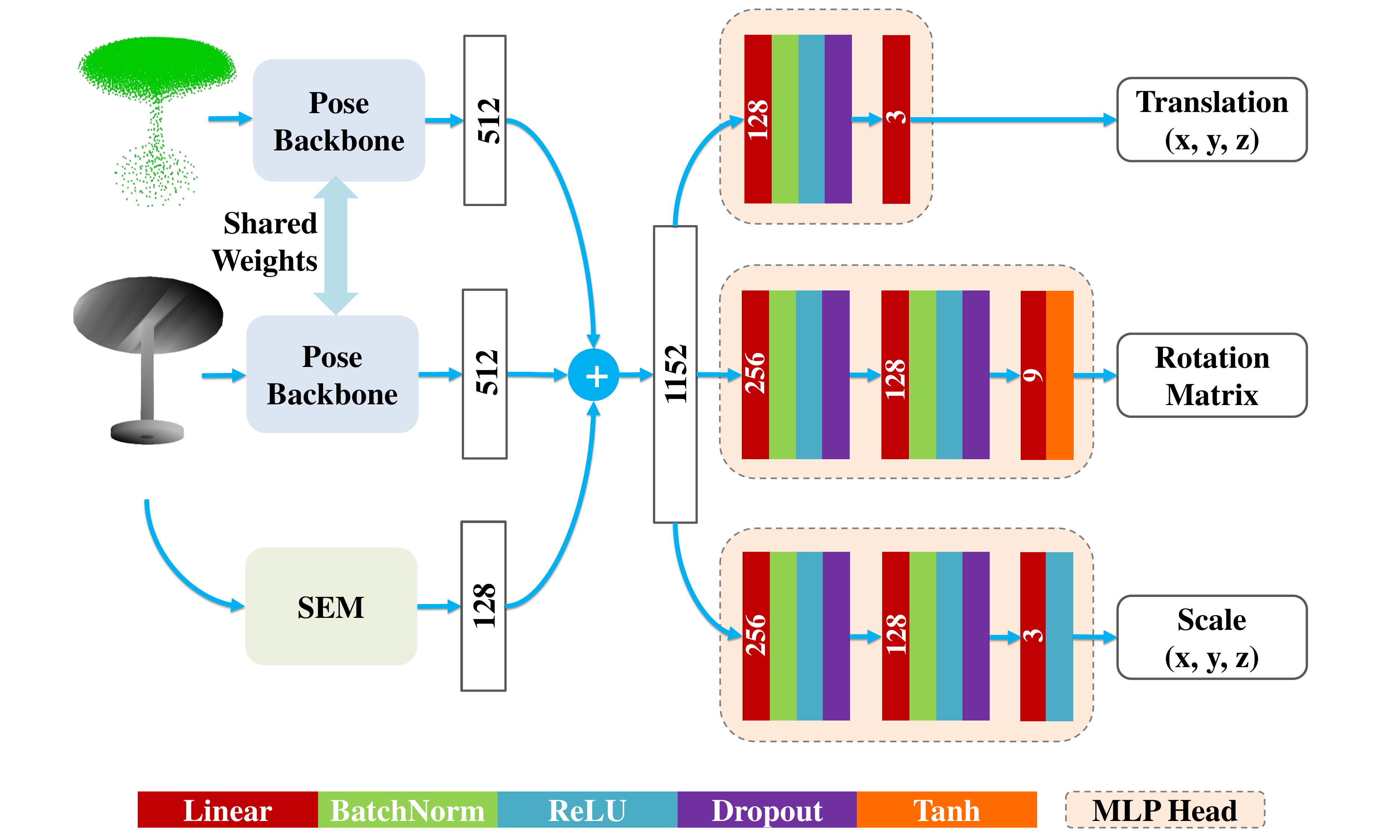}
    \caption{Proposed architecture of OCAM using MinkowskiFCNN\cite{choy20194d} as backbone. We use a shared backbone for extracting pose features from CAD and completed object point clouds. The extracted features are concatenated with the symmetry feature vector from SEM and forwarded to 3 different MLP heads, one each for estimating translation, rotation and scale.}
    \label{fig:arch}
\end{figure}

\subsection{Symmetry Encoder Module (SEM)}
Ground truth labels for learning object-CAD alignment do not account for object symmetry, which can significantly impact the learning process, particularly for rotation and scale-related features. Since symmetry is defined along specific axes and depends on the object’s orientation, symmetry-related features are not naturally learned when training to estimate alignment parameters.
To address these challenges, we introduce SEM, an encoder module designed to extract symmetry-aware features. This module is implemented using a Transformer-based network \cite{wu2024point} and is trained for binary symmetry classification—identifying whether an object has symmetry or not—using the Scan2CAD dataset. We also experimented with training the network on multiple symmetry classes (No symmetry, 2-fold, 4-fold, and infinite symmetry) from Scan2CAD to capture more nuanced features. However, this approach led to a significant decline in performance. It is important to note that symmetry-related features are extracted from the CAD point cloud, not the object point cloud, as the symmetry of an object with an arbitrary pose is ill-defined, and CAD models in synthetic datasets are in a common canonical pose. SEM is trained separately from OCAM, the alignment estimation module, and the embedding vector from SEM is used for training and inference in OCAM. Training details for SEM can be found in \ref{sem-exp}

\vspace{-2mm}
\subsection{Object-CAD Alignment Module (OCAM)}
Our method for learning object-CAD alignment utilizes a Siamese-style network that processes pairs of object and CAD point clouds to estimate the alignment parameters between them. This approach is implemented in OCAM, which consists of a 3D CNN backbone \cite{choy20194d} for feature extraction, followed by separate Multi-Layer Perceptron (MLP) heads to regress translation, rotation, and scale parameters. The output features from the backbone network are concatenated with the symmetry vector generated by SEM. The combined features are then processed through each MLP head to predict the translation ($t_x, t_y, t_z$), which is formulated as an offset from the centroid of the object point cloud, as well as the rotation (3x3 rotation matrix) and scale ($s_x, s_y, s_z$) parameters. An overview of this process is illustrated in Figure \ref{fig:arch}.

We further support the alignment training process by incorporating Chamfer Loss \cite{lin_infocd_2023} as a symmetry-aware loss function. Chamfer Loss is defined as follows:
\begin{align}
Loss_{cl} = \frac{1}{N_1} \sum_{o \in O} \min_{g \in GT} \|o - g\|_2^2 + \frac{1}{N_2} \sum_{g \in GT} \min_{o \in O} \|o - g\|_2^2    
    \label{chamfer_loss}
\end{align}
where O is the CAD point cloud (with N1 points) transformed using the 4x4 transformation matrix formed by the predicted alignment parameters and GT is the object point cloud with N2 points.
In addition, we also utilize weighted $L_1$ loss as follows:
\begin{align}
    L_{1\_pose} = \lambda_t L_1(t_{gt}, t_{p}) + \lambda_r L_1(r_{gt}, r_{p}) + \lambda_s L_1(s_{gt}, s_{p})
    \label{weighted_l1_loss}
\end{align}
where ($t_{gt}, r_{gt}, s_{gt}$) are ground truth translation, rotation and scale parameters, ($t_{p}, r_{p}, s_{p}$) are predicted parameters, and ($\lambda_t, \lambda_r, \lambda_s$) are their corresponding loss weights. Based on empirical studies, we found values of $\lambda_t=2, \lambda_r=3, \lambda_s=2$ to work best.
Our final loss formulation is as follows:
\begin{align}
    loss = 5\times Loss_{cl} + L_{1\_pose}
    \label{final_loss}
\end{align}
\vspace{-4mm}
\section{Dataset}
\label{sec-datasets}
To train and validate our approach, particularly CAPCM, we require datasets tailored for point cloud completion. As discussed in \ref{related-pccompletion}, real-world datasets for this task are scarce, while modern deep learning benefits from large-scale data. Moreover, existing synthetic datasets often differ greatly from real-world scans, hindering generalization. To bridge this gap, we introduce two datasets—S2C-Completion and SN-Indoor—designed for benchmarking on real-world data and reducing the domain gap between synthetic and real-world indoor scenes. The following subsections detail each dataset.

\vspace{-2mm}
\subsection{S2C-Completion dataset}
\label{S2C-Completion-dataset}
To address the lack of real-world point cloud completion datasets, and the limitations of Scan2CAD in this context (as discussed in \ref{sec-datasets}), we introduce S2C-Completion, an expert annotated dataset that combines instance annotations from ScanNet with pose and CAD model information from Scan2CAD and ShapeNet. 
S2C-Completion prioritizes precise alignment and considers finer geometric details to ensure an accurate match between scan instances and corresponding CAD models, resulting in a high-quality dataset for real-world indoor point cloud completion and object-CAD alignment tasks. A few samples from the dataset are shown in Figure \ref{fig:s2ccomp} (additional examples provided in supplementary materials), with CAD models aligned with instance point clouds and bounding boxes for visualizing pose. The full dataset is available \href{https://github.com/kumarhiranya/S2CCompletion}{here}. S2C-Completion consists of 8,535 samples, with 6,671 allocated to the training set and 1,864 to the test set. For each CAD model annotation from Scan2CAD, we provide a key ($scannet\_instance\_id$) that explicitly maps the CAD model to a ground truth object instance in ScanNet via its instance ID. If no suitable match is found, the key value is -1; otherwise, it is the same as the instance label ID of the matched object.
\begin{figure}[htbp]
    \centering
    \includegraphics[width=\columnwidth]{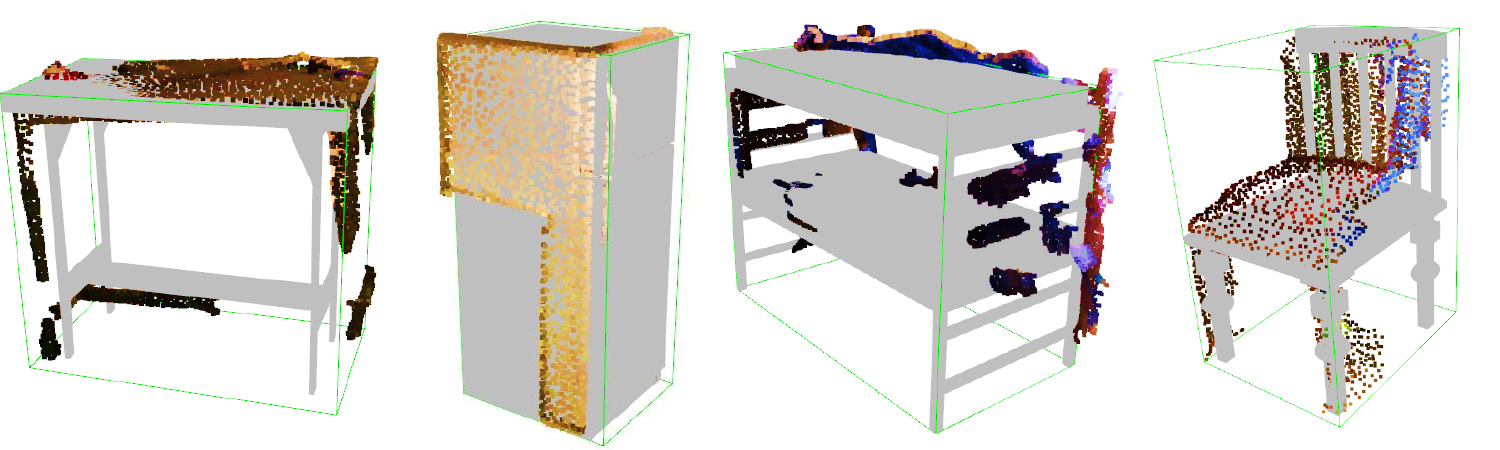}
    \caption{Annotations from S2C-Completion dataset with CAD model (grey) and corresponding ScanNet ground truth instance point clouds. Pose of the object is visualized as a green bounding box around the object.}
    \label{fig:s2ccomp}
\end{figure}
\vspace{-2mm}
\subsection{SN-Indoor}
\label{shapenet-indoor}
\begin{figure}[htbp]
    \centering
    \includegraphics[width=.95\columnwidth]{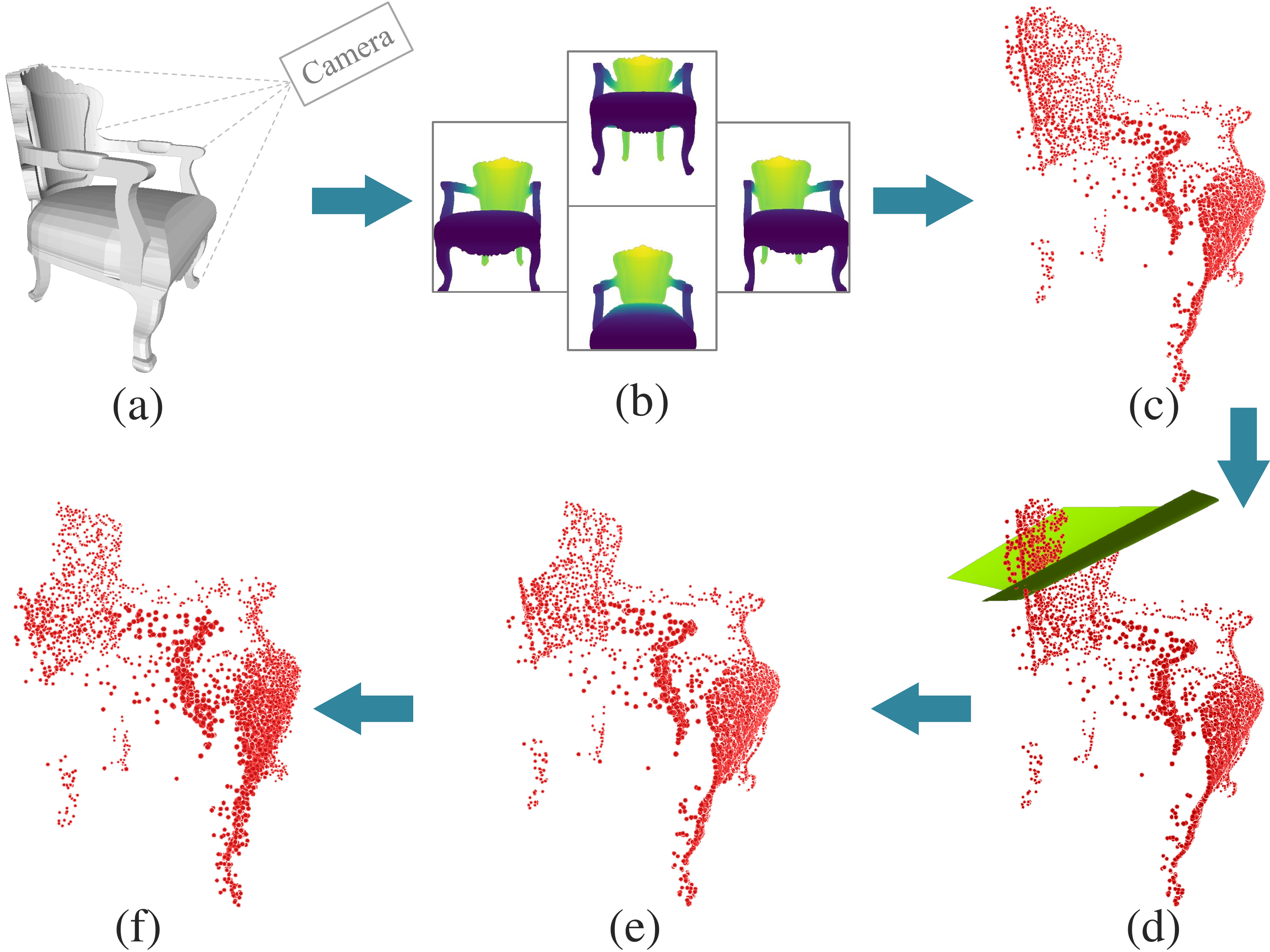}
    \caption{Incomplete point cloud generation steps from synthetic mesh. (a) Synthetic mesh. (b) Multi-view depth maps generated through single perspective ray-casting. (c) Occluded point cloud based on camera perspective. (d) Randomly generated non-linear plane. (e) Cropped point cloud. (f) Final output after adding Gaussian noise.}
    \label{fig:incomplete_pcd_generation}
\end{figure}
To generate synthetic data from ShapeNet, we begin by selecting a 3D model (Figure \ref{fig:incomplete_pcd_generation}(a)) and positioning a virtual camera at a random viewpoint around the object. Using ray-casting, we capture a single-perspective point cloud, which, unlike traditional 2D projection methods used in some synthetic datasets, better emulates real-world scenarios by reducing sampling density with distance, resulting in non-uniform surface coverage.  
Next, we slightly translate the camera in both vertical and horizontal directions (Figure \ref{fig:incomplete_pcd_generation}(b)) without significantly altering the perspective. The process is repeated, and the resulting point clouds are merged (Figure \ref{fig:incomplete_pcd_generation}(c)), mimicking camera movements during real-world scanning and generating a more comprehensive 3D point cloud instead of a predominantly planar one.  
To simulate occlusions, common in cluttered indoor environments, we crop a portion of the point cloud using a randomly generated non-linear plane (Figure \ref{fig:incomplete_pcd_generation}(d,e)), as detailed in the supplementary materials. We repeat this process multiple times, depending on the difficulty mode, to simulate multi-object occlusion. This approach more accurately represents real-world occlusions compared to linear plane-based cropping. The proportion of cropped points are determined by the dataset difficulty setting—easy (25\%), medium (50\%), and hard (75\%). Finally, a small amount of Gaussian noise is added to the point cloud to simulate sensor imperfections (Figure \ref{fig:incomplete_pcd_generation}(e)). We provide further details on the data augmentation techniques in the supplementary materials.
\vspace{-3mm}
\section{Experiments}

\begin{table*}[htbp]
    \centering
        \begin{tabular}{|l|c|c|c|c|c|c|c|c|c|}
            \hline
              & bath & bookshelf & cabinet & chair & other & sofa & table & Class avg & Avg \\
            \hline
            FPFH \cite{rusu2009fpfh} & 0.00 & 1.92 & 0.00 & 10.00 & 5.41 & 2.04 & 1.75 & 2.57 & 4.45 \\
            SHOT \cite{tombari2010shot} & 0.00 & 1.43 & 1.16 & 7.08 & 3.57 & 1.47 & 0.44 & 1.83 & 3.14 \\
            Li et al. \cite{li2015database} & 0.85 & 0.95 & 1.17 & 14.08 & 6.25 & 2.95 & 1.32 & 4.38 & 6.03 \\
            3D Match \cite{zeng_3dmatch_2017} & 0.00 & 5.67 & 2.86 & 21.25 & 10.91 & 6.98 & 3.62 & 6.48 & 10.29 \\
            Scan2CAD \cite{avetisyan_scan2cad_2019} & 36.2 & 36.4 & 34 & 44.26 & \textbf{70.63} & 30.66 & 30.11 & 35.64 & 31.68 \\
            End-to-End \cite{avetisyan_end--end_2019} & 38.89 & 41.46 & 51.52 & 73.04 & 26.83 & 76.92 & 48.15 & 51.44 & 50.72 \\
            CIS2VR \cite{kumar2024cis2vr} & 49.66 & 19.52 & 29.92 & 67.47 & - & 54.02 & 56.54 & 46.19 & 60.25\\
            SceneCAD \cite{avetisyan_scenecad_2020} & 42.42 & 36.84 & 58.33 & 81.23 & 40.24 & \textbf{82.86} & 45.60 & 52.27 & 61.24 \\
            \hline
            CAOA (No Completion) - Sph & 55.75 & 45.50 & 45.50 & 74.29 & 38.31 & 63.70 & 53.93 & 48.63 & 60.92 \\
            CAOA (w/ CAPCM) - Sph       & 84.24 & 59.49 & 70.62 & 83.45 & 58.47 & 81.03 & 74.55 & 67.45 & 75.83 \\
            CAOA (w/ CAPCM+SEM) - SG    & 72.20 & 59.81 & 69.28 & 83.75 & 60.11 & 77.42 & 76.51 & 66.84 & 76.04 \\
            CAOA (w/ CAPCM+SEM) - Sph   & \textbf{86.51} & \textbf{61.21} & \textbf{72.23} & \textbf{84.52} & 60.55 & 81.65 & \textbf{78.26} & \textbf{69.17} & \textbf{77.51} \\
            CAOA (w/ CAPCM+SEM) - GT   & 88.71 & 73.58 & 81.35 & 93.87 & 75.00 & 87.14 & 84.62 & 78.15 & 86.54 \\
            
            
            \hline
        \end{tabular}
    \caption{Alignment results for various methods on the Scan2CAD\cite{avetisyan_scan2cad_2019} benchmark. Numbers represent alignment accuracy for each category, higher is better. Last five rows show results on various configurations of CAOA, with the last row showing CAOA's performance on ground truth (GT) ScanNetv2 labels.}
\vspace{-0.05in}
\label{tab:alignment_results}
\end{table*}

\vspace{-2mm}
\subsection{Point Cloud Completion}
\label{exp-odg-datasets}
To benchmark the generalizability of synthetic datasets on real-world data, we use the algorithm proposed by Cai et al. \cite{cai2024odg}, ODGNet, which currently ranks as the top point cloud completion method on ShapeNet. We train on ShapeNet-derived synthetic sets and benchmark on real-world data from S2C-Completion.

\vspace{-4mm}
\paragraph{Training}
We use AdamW (lr 5$e^{-4}$, weight decay) with LambdaLR (step 20, $\gamma{=}0.8$, min 5$e^{-6}$), multi-level Chamfer Distance L1 (CDL1) loss, and orthogonal constraints on all learnable dictionaries. We train for 200 epochs or until convergence with a batch size of 32, which takes $\sim$1.5 days on one Nvidia RTX A6000 GPU.

\vspace{-4mm}
\paragraph{Results}
Table~\ref{tab:odg_performance} reports Chamfer Distance L1/L2 (CDL1/CDL2) on S2C-Completion for ODGNet trained on PCN \cite{yuan2018pcn}, ShapeNet-34/55 \cite{yu2021pointr}, and SN-Indoor; we also train on real S2C-Completion (S2C-C), a mix (SN-Indoor+S2C-C), and the mix with 100\,cm context (SN-Indoor+S2C-C+Ctxt\_100). Our results show a significant improvement in generalization when using the proposed SN-Indoor synthetic dataset, highlighting the effectiveness of our augmentation techniques. When training exclusively on the real-world data from S2C-Completion, performance plateaus after 80 epochs. However, augmenting this data with synthetic datasets prevents this stagnation, enabling the model to converge further and achieve better results than training on either dataset alone. Finally, incorporating context data into the combined dataset improves the model’s performance, underscoring the value of contextual information for completion. Notably, since SN-Indoor lacks context data, the algorithm was trained on mixed samples, with the S2C-Completion samples containing context, while the others did not. We also explored injecting explicit global and local semantic features while training CAPCM, but we found no improvements, suggesting that the algorithm might be learning similar features implicitly while training for completion.

\begin{table}[htpb]
\centering
\begin{tabular}{| p{4.5cm}| p{1.5cm}|c|c|c|}

\hline
\textbf{Train Dataset} & \textbf{CDL1}$\downarrow$ & \textbf{CDL2}$\downarrow$ \\
\hline
PCN\cite{yuan2018pcn} & 161.521 & 137.124 \\
\hline
ShapeNet-55\cite{yu2021pointr} & 94.103 & 78.029 \\
\hline
ShapeNet-34\cite{yu2021pointr} & 94.014 & 77.893 \\
\hline
SN-Indoor (ours) & 51.193 & 29.662 \\
\hline
S2C-Completion (ours) & 46.802 & 25.692 \\
\hline
SN-Indoor + S2C-C & 34.986 & 4.999 \\
\hline
SN-Indoor + S2C-C + Ctxt\_100 & \textbf{22.471} & \textbf{2.3} \\
\hline

\end{tabular}
\caption{Performance metrics of ODGNet (CD-L1$\times10^{-3}$, CD-L2$\times10^{-3}$), lower is better, trained on various datasets and evaluated on the S2C-Completion benchmark.}
\label{tab:odg_performance}
\end{table}

\vspace{-3mm}
\subsection{CAD Alignment}
\label{cad-alignment-exp}
We evaluate CAOA using the benchmarks defined by authors in \cite{avetisyan_scan2cad_2019}, considering an alignment accurate if the translation, rotation, and scale errors are within 20 cm, $20^\circ$, and 20\% of the ground truth, respectively. Since CAOA operates as a modular framework, taking input from an instance segmentation algorithm, we integrate a recent instance segmentation approach proposed by Shin et al. \cite{shin2024spherical}. 

\vspace{-3mm}
\paragraph{Training}
The network is trained using the loss function defined in Equation \eqref{final_loss} and optimized with AdamW. The training process employs an initial learning rate of $1e^{-4}$, weight decay of $1e^{-4}$, and a Cosine Annealing scheduler that lowers the minimum learning rate to $1e^{-6}$ over the training period. Training runs for 150 epochs or until convergence, taking approximately 2 hours on an RTX 3090.

\vspace{-4mm}
\paragraph{Results}
Table \ref{tab:alignment_results} presents the performance of CAOA on the Scan2CAD alignment benchmark. To ensure a fair comparison with existing modular methods, we also evaluate CAOA using the same instance segmentation algorithm (SoftGroup \cite{vu2022softgroup}) as used in CIS2VR \cite{kumar2024cis2vr}. Our evaluation is conducted on ScanNet's validation set, consisting of 312 scenes. The findings indicate that CAOA improves class average accuracy by approximately 17\% and overall accuracy by around 16\%, significantly surpassing the performance of existing methods and validating the effectiveness of our proposed approach. 
{Lastly, we evaluate CAOA on ground truth ScanNetv2 labels to assess the impact of instance segmentation errors. Results suggest \(\sim10\%\) alignment performance loss due to segmentation errors, indicating potential gains from improving instance point cloud quality.}
For qualitative comparison, we provide examples in the supplementary materials.

\vspace{-2mm}
\subsection{Symmetry Encoding}
\label{sem-exp}
Our SEM is trained on ShapeNet data, incorporating symmetry annotations from Scan2CAD. The architecture uses PointTransformerV3 \cite{wu2024point} as the feature extraction backbone, with an embedding dimension of 128 and an MLP head that employs Softmax activation for the final classification.
The model is trained using Binary Cross Entropy (BCE) loss and the AdamW optimizer, with a learning rate initialized at $1e^{-4}$, weight decay of $1e^{-3}$, and a Cosine Annealing scheduler that reduces the minimum learning rate to $1e^{-6}$. Training lasts for 150 epochs or until convergence, requiring approximately one hour on an RTX 3090. The network achieves 95.8\% accuracy and a 94.1\% F1-Score, demonstrating its successful learning of symmetry-related features.

\subsection{Ablation studies}
We conduct ablation studies to assess the impact of different CAOA components on the final results. In these studies, we train and evaluate CAOA with various components disabled and present the results in Table \ref{tab:alignment_results}.

\vspace{-5mm}
\paragraph{Effect of Point Cloud Completion}
The entries in Table \ref{tab:alignment_results} labeled “CAOA (No Completion) - Sph” and “CAOA (w/ CAPCM) - Sph” present the results of running our algorithm without point cloud completion (no CAPCM) and with CAPCM, respectively. The results demonstrate that CAPCM has a significant impact on performance, leading to improvements of over 40\% in certain categories. Furthermore, the overall and weighted average performance increase by 15\% and 19\%, respectively, when CAPCM is included, highlighting its importance for achieving better results.

\vspace{-5mm}
\paragraph{Effects of Training CAPCM on Synthetic Datasets}
We also investigate the impact of training CAPCM on different datasets on the final alignment performance of CAOA. The results, included in supplementary materials, indicate that the choice of dataset for CAPCM training significantly influences alignment performance. When trained on the ShapeNet-55/34 and SN-Indoor datasets, overall performance decreases by about 7\% compared to our default training configuration (using a mix of ScanNet-Indoor and S2C-Completion with 100 cm context), suggesting that these datasets generalize well to real-world settings, but don't close the gap completely. Notably, minor variations in point cloud completion performance between ShapeNet-55/34 and SN-Indoor do not lead to noticeable differences in alignment results. However, when trained on the PCN dataset, performance drops dramatically by approximately 23\%, even performing worse than training with incomplete point clouds, highlighting the importance of proper dataset selection for training CAPCM.

\vspace{-5mm}
\paragraph{Effects of Symmetry Features}
The rows labeled “CAOA (w/ CAPCM) - Sph” and “CAOA (w/ CAPCM+SEM) - Sph” in Table \ref{tab:alignment_results} show the results of training CAOA without and with symmetry information (SEM), respectively. Including SEM in the training process leads to an approximate 2\% improvement, demonstrating that symmetry features contribute positively to alignment estimation. Note that Chamfer Loss is only used while training with SEM, as it doesn't lead to any noticeable improvements otherwise.



\subsection{Runtime Analysis}
We assess CAOA's runtime by executing the entire pipeline—from 3D instance segmentation to final alignment prediction—on scenes with varying object counts. The evaluation is conducted on a system with an AMD Ryzen 5900X processor and an Nvidia RTX 3090 GPU. Table \ref{tab:runtime-analysis} compares CAOA's runtime with various existing methods, showing results for both SoftGroup and SphericalMask as instance segmentation approaches. The results indicate that our method outperforms existing methods in runtime efficiency, averaging approximately 0.58 seconds per scene across the ScanNet validation dataset. However, it is important to note that the results provided by CIS2VR include several other steps such as scene reconstruction in Unity. A detailed runtime analysis of individual pipeline modules reveals that the point cloud completion module requires around 10.3 ms, while the relative pose estimation takes 3.5 ms per object, with a nearly linear increase as the number of objects rises. The low inference time makes CAOA suitable for dynamic or real-time applications.

\begin{table}[htbp]
    \centering
        \begin{tabular}{| p{2.5cm} | l | l | l |}
            \hline
              & 7 Objects & 16 Objects & 20 Objects \\
            \hline
            Scan2CAD \cite{avetisyan_scan2cad_2019} & 288.60s & 565.86s & 740.34s \\
            \hline
            SceneCAD \cite{avetisyan_scenecad_2020} & 2.0s(5) & - & 2.60s(26) \\
            \hline
            End-to-End \cite{avetisyan_end--end_2019} & 0.62s & 1.11s & 2.60s \\
            \hline
            CIS2VR \cite{kumar2024cis2vr} & 0.55s & 0.61s & 0.66s \\
            \hline
            CAOA + SG & 0.38s & 0.54s & 0.59s \\
            \hline
            CAOA + Sph & 0.35s & 0.5s & 0.56s \\
            
            \hline
        \end{tabular}
    \caption{Runtime (in seconds) comparison with existing methods for scenes with varying number of objects. Note that evaluation on a matching number of objects wasn't available for SceneCAD, hence we mention the number of objects in parenthesis.}
\vspace{-0.05in}
\label{tab:runtime-analysis}
\end{table}
\vspace{-4mm}
\section{Limitations and Future Work}
\label{sec:limitations}
\vspace{-2mm}
CAOA achieves state-of-the-art performance in object–CAD alignment but still offers room for enhancement. At present, the point cloud completion and pose estimation modules are trained separately; a unified training strategy could allow the completion process to be guided directly by pose estimation objectives. Moreover, incorporating a context-aware pose estimation module that accounts for surrounding objects and scene structures—similar to the approach in \cite{avetisyan_scenecad_2020}—could further improve accuracy.  
While CAOA is designed for object–CAD alignment, the underlying techniques can also benefit other stages of semantic reconstruction, including CAD retrieval and layout estimation.

\vspace{-2mm}
\section{Conclusion}
\vspace{-2mm}
\label{sec:conclusion}
CAOA is a single-object scan–based object–CAD alignment method that leverages context-aware point cloud completion and symmetry encoders prior to pose estimation, yielding cleaner, more complete inputs and symmetry-aware features that enhance alignment accuracy. We introduce S2C-Completion, an expert-annotated real-world completion dataset extending Scan2CAD, and SN-Indoor, a synthetic data generation technique tailored to indoor environments, both enabling improved training and benchmarking for completion algorithms. We further propose a symmetry-aware learning process for more robust pose feature extraction.

\paragraph{Acknowledgments}
This material is based upon work supported by the National Science Foundation (NSF) under Grant Nos. 2603236 and 2532731. Any opinions, findings, and conclusions or recommendations expressed in this material are those of the author(s) and do not necessarily reflect the views of the NSF.

{
    \small
    \bibliographystyle{ieeenat_fullname}
    \bibliography{main}
}

\newpage
\section{Appendix Section}
\label{sec:appendix_section}

\begin{figure*}[ht]
    \centering
    \includegraphics[width=\textwidth]{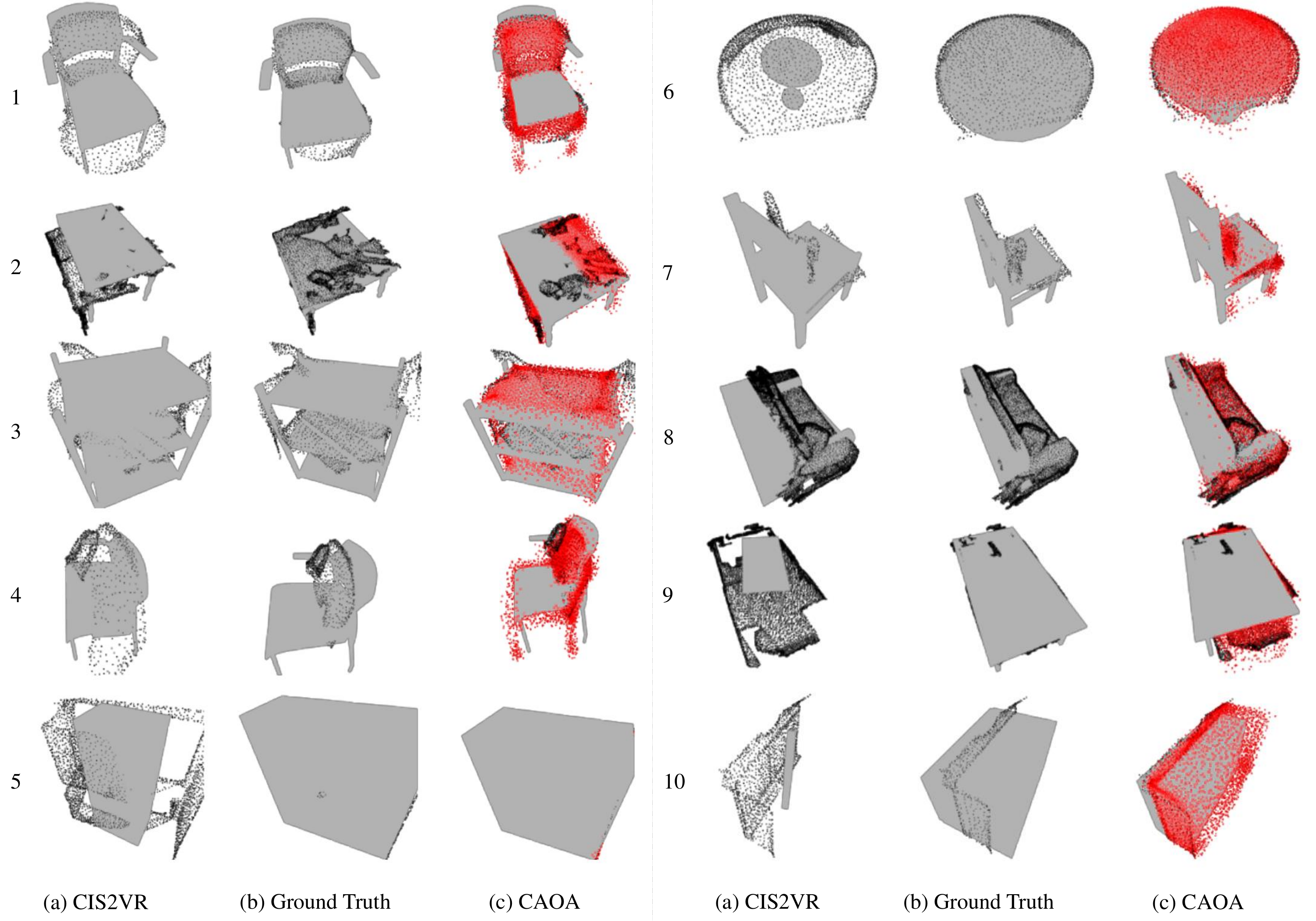}
    \caption{A qualitative comparison of alignment results using CIS2VR\cite{kumar2024cis2vr}, ground truth and CAOA. Incomplete instance point clouds are shown in black, and completed point clouds in red. Predicted poses are in gray.}
    \label{fig:cis2vr-comparison1}
\end{figure*}
\subsection{Qualitative Analysis}
Figure \ref{fig:cis2vr-comparison1} presents qualitative CAD alignment results, comparing CAOA (c) and CIS2VR \cite{kumar2024cis2vr} (a) with ground truth annotations (b). The input object point clouds are instance predictions generated by SoftGroup \cite{vu2022softgroup} for indoor scenes from the ScanNet dataset \cite{dai2017scannet}. For each set of visualizations, CIS2VR results are shown on the left (a), ground truth (GT) in the middle (b), and CAOA results on the right (c). Object point clouds are depicted in black, completed point clouds (for CAOA) are in red, and CAD objects are aligned using the predicted pose in gray. The figures demonstrate a significant improvement in alignment performance with CAOA, particularly in scenarios involving incomplete instance point clouds. These results underscore the advancements achieved through the proposed approach.

\begin{figure*}[htbp]
    \centering
    \includegraphics[width=\textwidth]{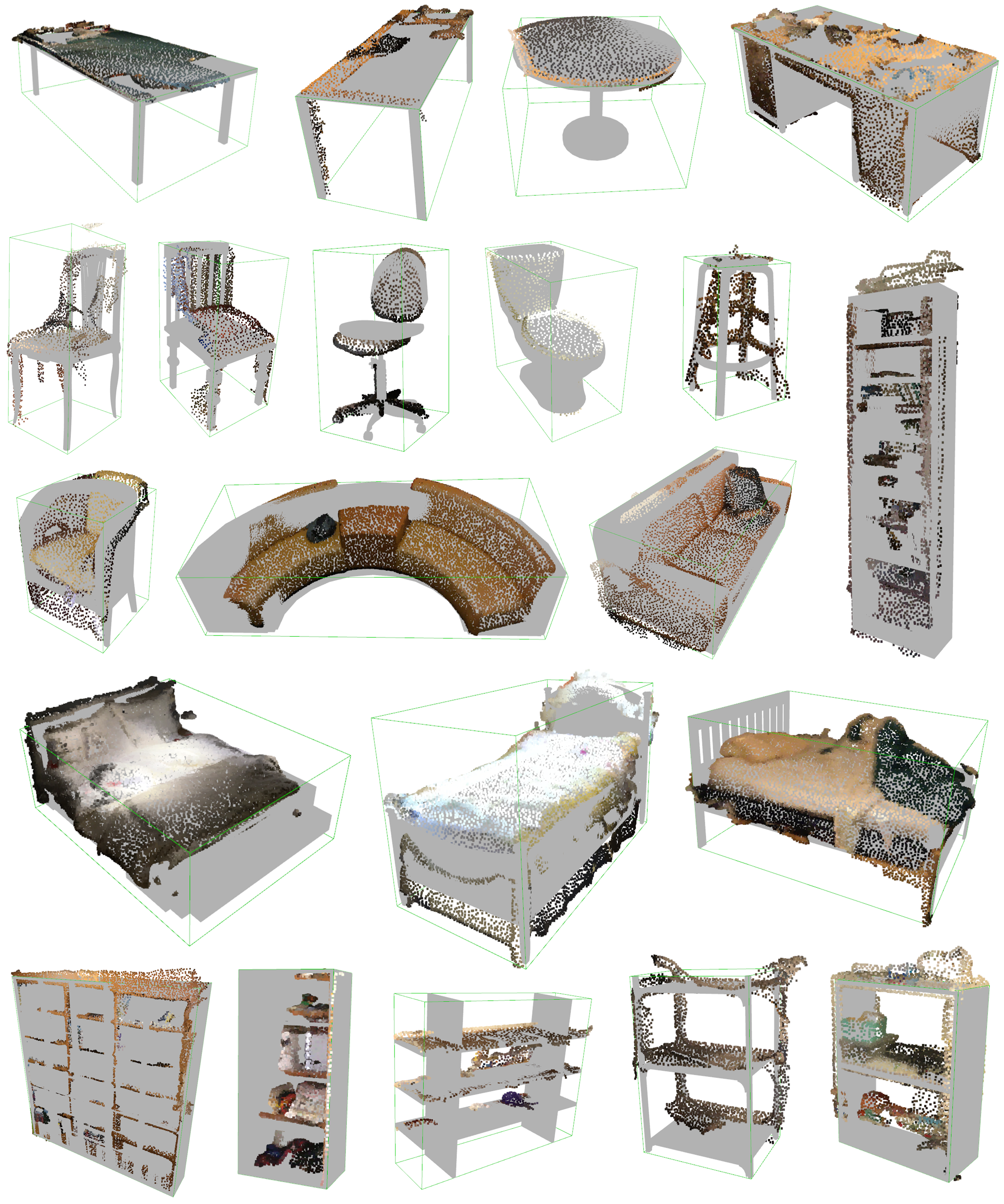}
    \caption{Examples from S2C-Completion dataset showing colored scan instance point clouds and aligned CAD models (gray). Each object's 9-DoF pose is visualized as a green bounding box.}
    \label{fig:S2C-Completion-supp}
\end{figure*}

\subsection{S2C-Completion Dataset}
We provide further examples from the S2C-Completion dataset in Figure \ref{fig:S2C-Completion-supp}. The figures show scan object color point clouds with aligned CAD models in gray. The single object pose of each object instance is visualized using a green bounding box surrounding each object.

\subsection{Alignment Performance on Synthetic Datasets}
\begin{table}[ht]
\centering
\begin{tabular}{|p{3.5cm}|p{1cm}|p{2.5cm}|}
\hline
\textbf{CPCM Train} & \textbf{Avg}$\uparrow$ & \textbf{Weighted Avg}$\uparrow$ \\
\hline
PCN                         & 55.68                           & 47.69 \\
ShapeNet-55/34              & 70.82                           & 63.77 \\
SN-Indoor             & 70.78                           & 63.89 \\
S2C-Completion             & 71.05                           & 64.38 \\
SN-Indoor + S2C-C + Ctxt 100 cm & 77.51                        & 69.17 \\
\hline
\end{tabular}

\caption{
Alignment performance of CAOA with CPCM trained on different datasets. We observe that the performance of CPCM on point cloud completion has significant impact on the final alignment performance.
}
\label{tab:cpcm_accuracy}
\end{table}

\subsection{SN-Indoor}
\begin{figure*}[htpb]
    \centering
    \includegraphics[width=0.95\textwidth]{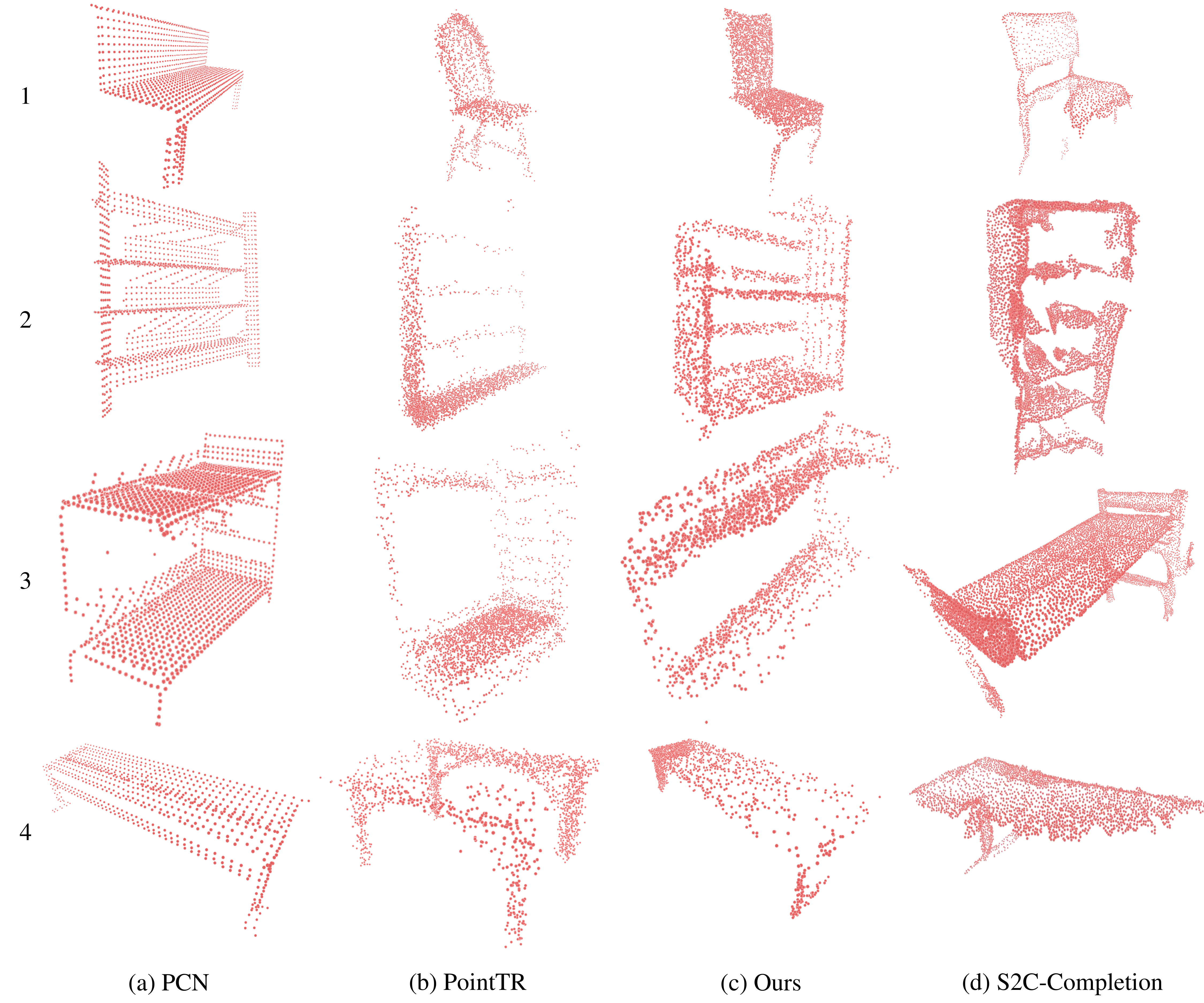}
    \caption{A qualitative comparison of synthetic point cloud completion datasets. The last column is from real-world scans.}
    \label{fig:comparison-supp}
\end{figure*}

Figure \ref{fig:comparison-supp} presents a comparative analysis of our dataset alongside widely used synthetic datasets and a real-world dataset. Real-world scans are captured using an RGB-D sensor, with the intrinsic and extrinsic parameters of the camera utilized to transform image coordinates into world coordinates. Multiple camera frames capture distinct sets of world coordinates, which are integrated to generate the final point cloud. 

\subsubsection{Perspective Point cloud Generation}
In our approach, we aim to replicate these real-world conditions while generating incomplete point clouds from synthetic mesh data. The following equations illustrate the transformation technique utilized in our approach:


\begin{align}
PCD_{o} = \sum_{n = 1}^N R_n^{-1} (d . K^{-1} (x, y) - t_n) 
    \label{eq:transform_pcd1}
\end{align}

Here, $PCD_{o}$ denotes the occluded point cloud obtained after ray casting. $n$ represents the number of frames captured from a single camera perspective. $K$ and $[R|t]$ correspond to the camera model parameters. $d$ is depth of the $(x, y)$ image coordinate. 

\subsubsection{Non-Linear Cropping}
We use the following equations to define the non-linear surface ($f$) used for cropping. $r_{1}...r_9$ are randomly generated values used to define the shape of the plane. $G_{\sigma}$ is the added gaussian noise, where ${\sigma}$ is set to $0.005$.

\begin{align}
    fs(x) = r_1 * \sin(r_2*x + r_3) \\
fc(y) = r_4 * \cos(r_5*y + r_6) \\
f(x,y,z) = fs(x) + fc(y) +r_7*x + r_8*y + r_9
    \label{eq:transform_pcd2} \\
PCD_{i} = PCD_{o} [f(PCD_{o})] + G_{\sigma}
    \label{eq:transform_pcd3}
\end{align}

Here,  $PCD_{i}$ represents the final incomplete point cloud after cropping and adding Gaussian noise.



\end{document}